\documentclass[letter, 10pt,conference]{ieeeconf}
\IEEEoverridecommandlockouts
\input{texdraw}
\usepackage{amsmath,amsfonts}
\usepackage{algorithmic}
\usepackage{algorithm}
\usepackage{psfrag,epsfig,color,hyperref}
\usepackage{subfig}

\font\bfmath=cmmib10
\mathchardef\Gamma="7100
\mathchardef\Delta="7101
\mathchardef\Theta="7102
\mathchardef\Lambda="7103
\mathchardef\Xi="7104
\mathchardef\Pi="7105
\mathchardef\Sigma="7106
\mathchardef\Upsilon="7107
\mathchardef\Phi="7108
\mathchardef\Psi="7109
\mathchardef\Omega="710A

\mathchardef\alpha="710B
\mathchardef\beta="710C
\mathchardef\gamma="710D
\mathchardef\delta="710E
\mathchardef\epsilon="710F
\mathchardef\zeta="7110
\mathchardef\eta="7111
\mathchardef\theta="7112
\mathchardef\iota="7113
\mathchardef\kappa="7114
\mathchardef\lambda="7115
\mathchardef\mu="7116
\mathchardef\nu="7117
\mathchardef\xi="7118
\mathchardef\pi="7119
\mathchardef\rho="711A
\mathchardef\sigma="711B
\mathchardef\tau="711C
\mathchardef\upsilon="711D
\mathchardef\phi="711E
\mathchardef\chi="711F
\mathchardef\psi="7120
\mathchardef\omega="7121
\mathchardef\epsilon="7122

\mathchardef\varepsilon="7122
\mathchardef\vartheta="7123
\mathchardef\varpi="7124
\mathchardef\varrho="7125
\mathchardef\varsigma="7126
\mathchardef\varphi="7127
\mathchardef\imath="717B
\mathchardef\jmath="717C

\def\bff{{\mbox{\boldmath $f$}}}

\def\bfp{{\mbox{\boldmath $p$}}}

\def\bfr{{\mbox{\boldmath $r$}}}

\def\bfv{{\mbox{\boldmath $v$}}}

\def\bfI{{\mbox{\boldmath $I$}}}
\def\bfJ{{\mbox{\boldmath $J$}}}

\def\bfN{{\mbox{\boldmath $N$}}}

\def\bfsigma{{\mbox{\boldmath $\sigma$}}}

\def\bfLambda{{\mbox{\boldmath $\Lambda$}}}

\def\norm#1{\left\|#1\right\|}

\def\smallbfW{{\raise1.5pt\hbox{\mbox{\boldmath $_W$}}}}
\def\t{^{\rm T}}
\let\ts=\thinspace

\def\Re{\mathbb{R}}

\def\mypsfrag#1#2#3#4#5{
        \begin{figure}[htp]
           \begin{center}
              {\leavevmode
                 {\includegraphics[width=#1truecm]{#2.eps}}
              }
           \end{center}
           \vspace{#3}
           \caption{#4}
           \label{#5}
        \end{figure}
}

\def\my4psfrag#1#2#3#4#5#6#7#8{
        \begin{figure}[htp]
        \begin{center}
	        \begin{tabular}[h]{c c}
              {\leavevmode{\includegraphics[width=#1truecm]{#2.eps}}}
              &
              {\leavevmode{\includegraphics[width=#1truecm]{#3.eps}}} \\
              {\leavevmode{\includegraphics[width=#1truecm]{#4.eps}}}
              &
              {\leavevmode{\includegraphics[width=#1truecm]{#5.eps}}}
   	     \end{tabular}
           \vspace{#6}
           \caption{#7}
           \label{#8}
        \end{center}
        \end{figure}
}

\def\mydouble4psfrag#1#2#3#4#5#6#7#8{
        \begin{figure*}[htp]
        \begin{center}
            \begin{tabular}[h]{c c}
              {\leavevmode{\includegraphics[width=#1truecm]{#2.eps}}}
              &
              {\leavevmode{\includegraphics[width=#1truecm]{#3.eps}}} \\
              {\leavevmode{\includegraphics[width=#1truecm]{#4.eps}}}
              &
              {\leavevmode{\includegraphics[width=#1truecm]{#5.eps}}}
         \end{tabular}
           \vspace{#6}
           \caption{#7}
           \label{#8}
        \end{center}
        \end{figure*}
}

\def\mymatrix#1{ \left[\begin{matrix}#1\end{matrix}\right]}

\title{\LARGE \bf Connectivity maintenance by robotic Mobile Ad-hoc NETwork}
\author{Vaibhav Kumar Mehta, Filippo Arrichiello
\thanks{Vaibhav Kumar Mehta is a graduate student at the School of Engineering and Science of Jacobs University Bremen, Germany,
{\tt v.mehta@jacobs-university.de}}%
\thanks{F. Arrichiello is with the Department of Electrical and Information Engineering, University of Cassino and Southern Lazio, Via G. Di Biasio 43, 03043, Cassino (FR), Italy,
{\tt f.arrichiello@unicas.it}}
}

\begin{document}

\maketitle

\begin{abstract}

The problem of maintaining a wireless communication link between a fixed base station and an autonomous agent by means of a team of mobile robots is addressed in this work. Such problem can be of interest for search and rescue missions in post disaster scenario where the autonomous agent can be used for remote monitoring and first hand knowledge of the aftermath, while the mobile robots can be used to provide the agent the possibility to dynamically send its collected information to an external base station. To study the problem, a distributed multi-robot system with wifi communication capabilities has been developed and used to implement a Mobile Ad-hoc NETwork (MANET) to guarantee the required multi-hop communication. None of the robots of the team possess the knowledge of agent's movement, neither they hold a pre-assigned position in the ad-hoc network but they {\it adapt\/} with respect to the dynamic environmental situations. This adaptation only requires the robots to have the knowledge of their position and the possibility to exchange such information with their one-hop neighbours. Robots' motion is achieved by implementing a behavioural control, namely the Null-Space based Behavioural control, embedding the collective mission to achieve the required self-configuration. Validation of the approach is performed by means of demanding experimental tests involving five ground mobile robots capable of self localization and dynamic obstacle avoidance.

\end{abstract}

\section{Introduction}
Multi-robot systems have received an increasing attention in the last decades due to the advantages they present with respect to a single robot in terms of fault tolerance, flexibility in  mission execution, and possibility to use distributed sensing and actuation. The work~\cite{parker_springer08} presents an overview on multi-robot systems focusing on aspects like system architecture, communication, task allocation, and applications; while the work~\cite{parker2008distributed} deals with distributed intelligence issues and introduces a classification based on exhibited interactions. Recent researches in this field are mainly focused on distributed and cooperative robotic systems~\cite{bullo_princeton2008}, i.e., systems where each robot interacts only with its neighbours and, together, they generate the global behaviour of the team. The work~\cite{KumRusSuk_springer08} presents an overview on networked robots, that is, multiple robots that cooperate by network communication, and it explores the research challenges due to the interaction among control, communication and perception. 

In the rescue robotics scenario, the use of a single robot with a multitude of sensing capabilities has been studied and practised at large.  
Currently most of the robots in the search and rescue domain are human operated; while the need of well-trained operator is sometimes a requirement by the unpredictable conditions commonly found in disaster sites (collapsed buildings, gaps, holes, flooding ...), this limits the deployment of such robots readily. Even with the availability of autonomous ones, in situations where network capabilities are lacking the possible area to be explored is drastically reduced. Thus, the use of distributed networked robots make them very suitable for search and rescue scenarios, extending the range of communication and providing large scale surveillance at the same time. But dealing with networked robots poses the challenging problem of connectivity maintenance, i.e., performing autonomous missions while keeping the communication network formed by the robots/agents globally connected.

In reference~\cite{Pimentel} a group of mobile robots is required to maintain a wireless ad-hoc network by resorting to a distributed algorithm; each robot computes a first-order prediction of the network topology and, according to a proper cost function, estimates the position where the probability to maintain network connectivity is the largest. A behaviour-based approach is used in~\cite{Vazquez} to encourage a team of robots in maintaining a connected communication network during an exploration mission; the robots that form a bridge connection, i.e., a link whose removal disconnects the network, are forced to implement a connectivity-behaviour task. Reconfiguration aimed at implementing a fault-tolerant bi-connected configuration is proposed in~\cite{Basu}. Decentralized estimation procedure, based on graph theory,  are used in~\cite{yang2010decentralized} to allow each agent to track the algebraic connectivity of a time-varying graph; such measure of graph connectivity has been used in~\cite{sabattini2011decentralized} for the connectivity maintenance problem, and in~\cite{robuffo_RSS2011} in the framework of teleoperation of groups of UAVs (Unmanned Aerial vehicles).

In this paper we focus on the application where an autonomous agent, e.g., a robot or a human operator, navigates inside an area where communication infrastructure is lacking; even so, the agent is required to constantly communicate with a fixed base station. In the absence of other/stationary network devices, the explorable area would be limited by the maximum range of peer-to-peer wireless communication between the agent and the base station. Indeed, the presence of a set of autonomous mobile robotic devices that establish a wireless MANET and act as relays will be of significant importance to increase, via multi-hop, the maximum explorable area. The considered mobile devices are ground robots (aerial robots can be considered as-well) that are able to autonomously reconfigure without knowing in advance the motion of the leading agent. Each robot acts both as a host and a router, and it can directly communicate with robots/systems in its transmission range, i.e., its one-hop neighbours. The communication between the base station and the agent is thus achieved by data packets relayed over a sequence of intermediate nodes (the robots) using a store-and-forward multi-hop transmission principle.


Following the preliminary results in~\cite{AntArrChiSet_iros05,AntArrChiSet_ijmic06}, in this paper we propose a distributed control approach to solve the afore mentioned problem using a behaviour-based technique and a task activation strategy related to realistic characteristics of routing algorithms for mesh network. The proposed approach has been validated via challenging experimental tests in large indoor environment with a team of five ground mobile robots.
\begin{psfrags}
	\psfrag{base station}[][]{base station}
	\psfrag{mobile antennas}[][]{mobile antennas}
	\psfrag{agent}[][]{agent}
\begin{figure}
     \begin{center}
        \begin{tabular}[h]{c}
			\subfloat[]{\label{fig:mission1a}}
			\includegraphics[width=6truecm]{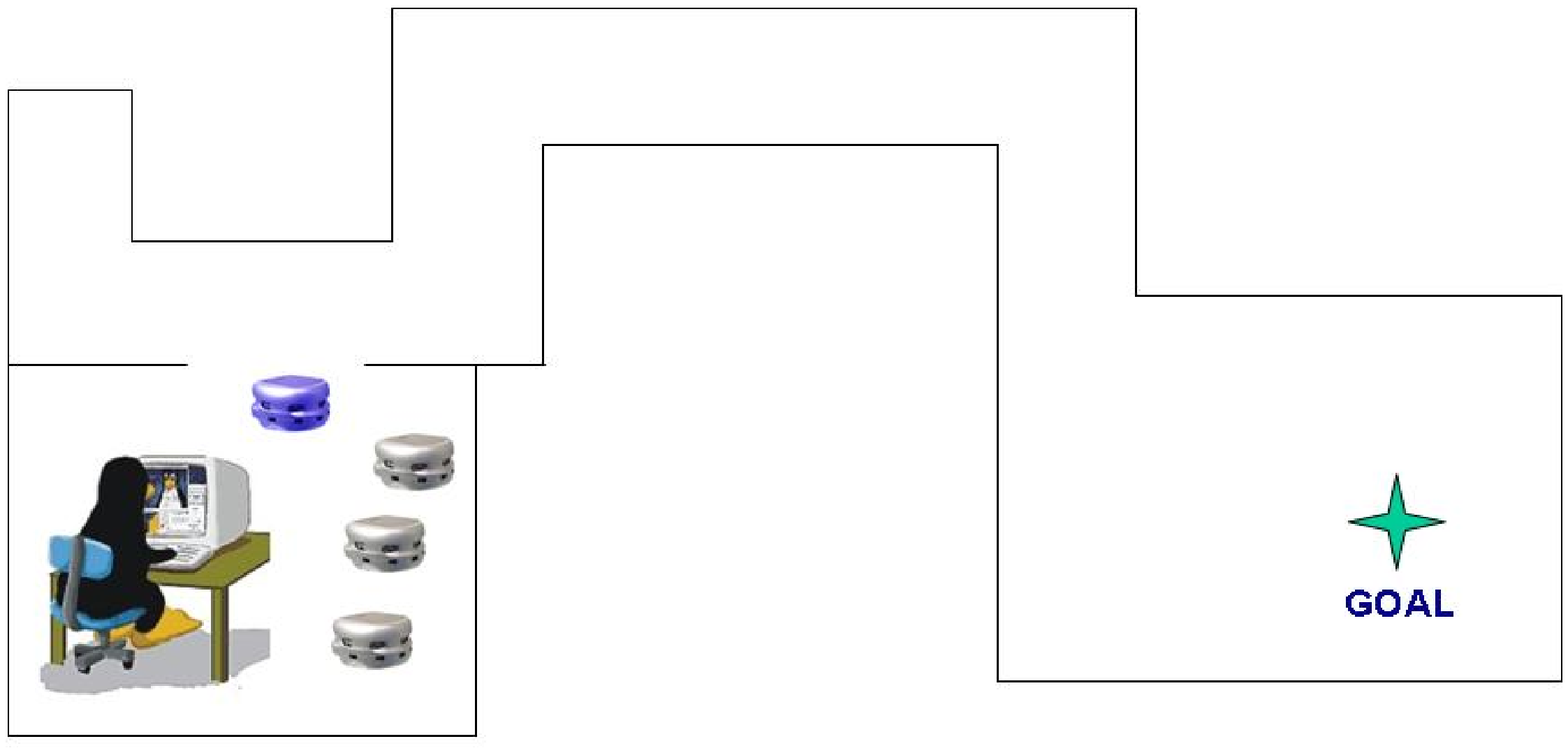}	\\
			\subfloat[]{\label{fig:mission1b}}
			\includegraphics[width=6truecm]{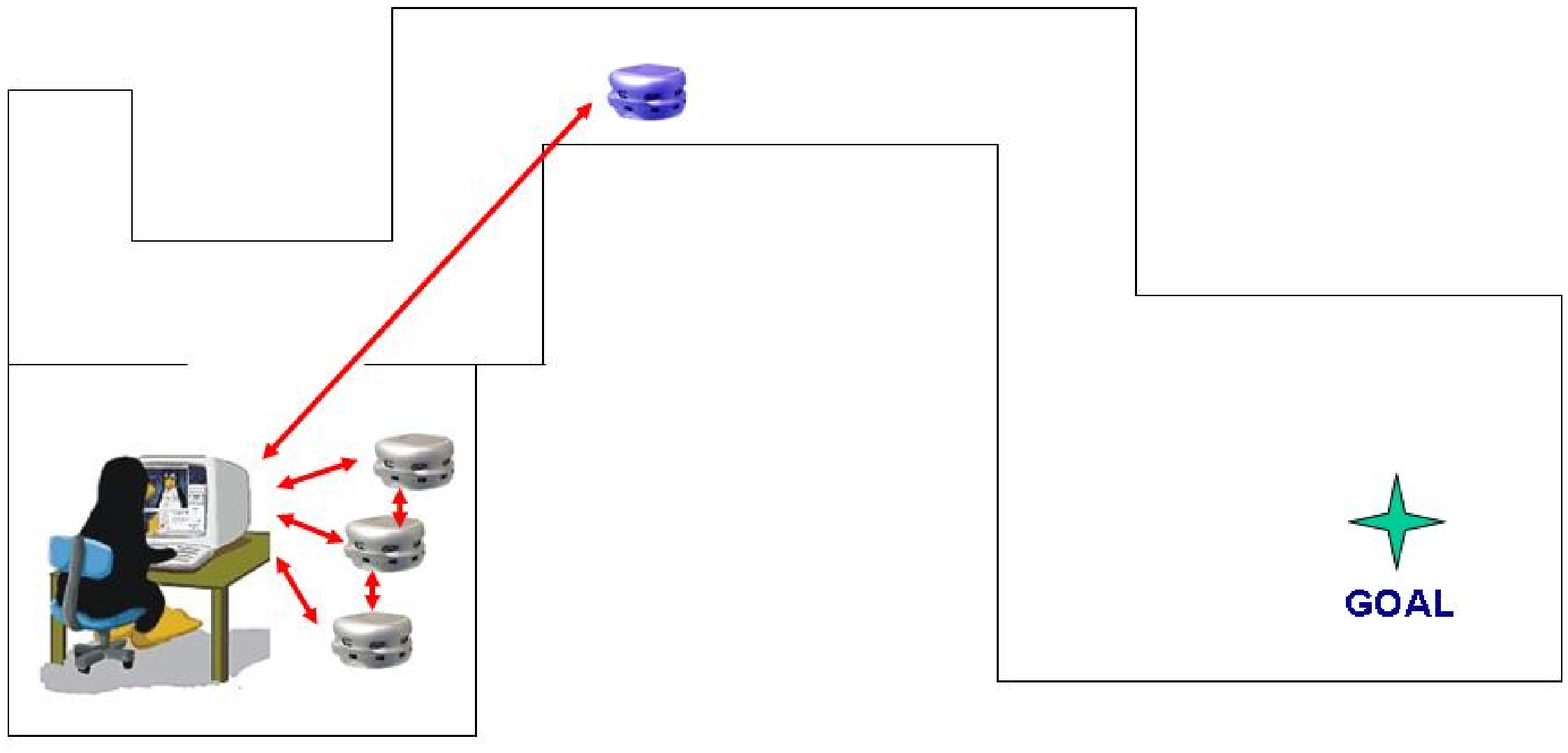}	\\
			\subfloat[]{\label{fig:mission1c}}
			\includegraphics[width=6truecm]{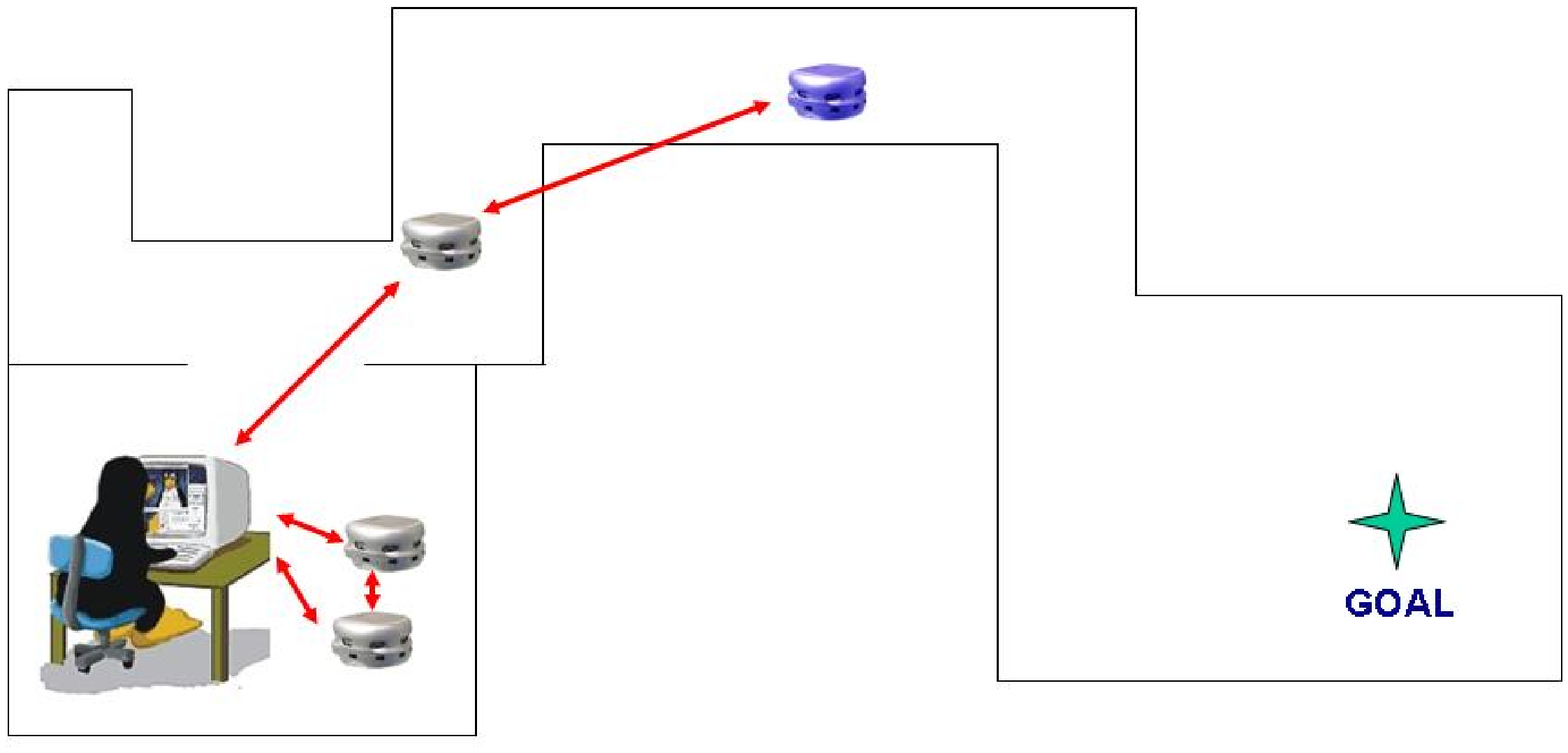}	\\
			\subfloat[]{\label{fig:mission1d}}
			\includegraphics[width=6truecm]{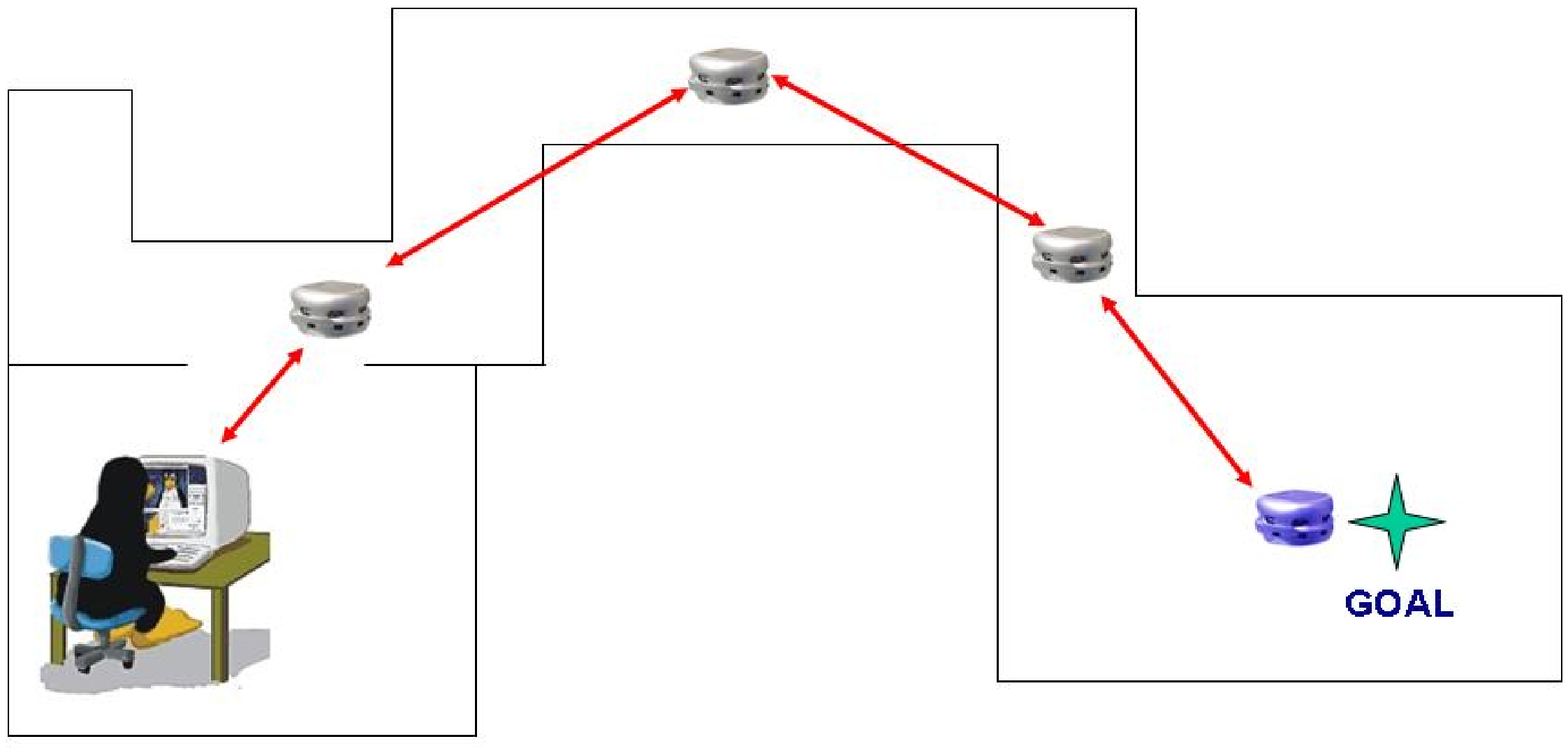}
        \end{tabular}   		
        \end{center}
        \caption{Sketch of the problem to be solved.}
        \label{fig:mission}
        \end{figure}
\end{psfrags}

\section{Connectivity maintenance strategy}\label{sec:connectivity}

Consider the scenario in Figure~\ref{fig:mission}, one autonomous agent (in blue) is moving according to a specific strategy that is unknown to the other devices. During the mission, the wireless communication connectivity between the agent and the fixed base station should be permanently guaranteed by a team of $n$~support mobile robots (in grey) that are required to realize an ad-hoc network and to increase the navigational coverage area of the agent. The maximum communication range between agent and base station is shown in Fig.~\ref{fig:mission1b}\ts.

We propose a distributed motion control strategy for the support robots to allow the autonomous agent accomplish its mission. Specifically, the support robots are required to simultaneously achieve multiple tasks like avoiding obstacles and collisions, generating a connected ad-hoc network for relaying the data from the lead agent to base station, and keeping the connectivity to the network. Moreover, they are required to use information from their on-board sensors and to exchange messages exclusively with their immediate neighbours in the established ad-hoc network.

The control policy is generated on the basis of the information available to the robot. In particular, we assume that each network node runs a routing protocol for ad-hoc/mesh networks, and that each robot has access to its dynamic routing table (any routing protocol is permissible). Thus, analysing the routing table, each robot can extract network layer information (ip address) about all other accessible robots inside the communication network (considering both one-hop and multi-hop), moreover, it can extrapolate what are the directly connected nodes (one-hop), and, for the non immediate neighbours, the first hop towards them. Figure~\ref{fig:routing} shows an example of routing table available to one of the robots in an assigned configuration; from the case in figure, robot~5 can infer from its routing table that it can directly communicate with robot~3,~4 and with the agent, and it knows that it can communicate with robots~1,~2 and the base station through robot 3 as the first hop.

In order to send messages to the base station, the autonomous agent elaborates information from its routing table and sends the message to the first hop towards the base. Iteratively, the robots on the path, that are the robots receiving the message originated by the agent (see Fig.~\ref{fig:path}), are required to forward it until it reaches the base station.

\begin{psfrags}
	\psfrag{base station}[][]{base station}
	\psfrag{mobile antennas}[][]{mobile antennas}
	\psfrag{agent}[][]{agent}
	\mypsfrag{6}{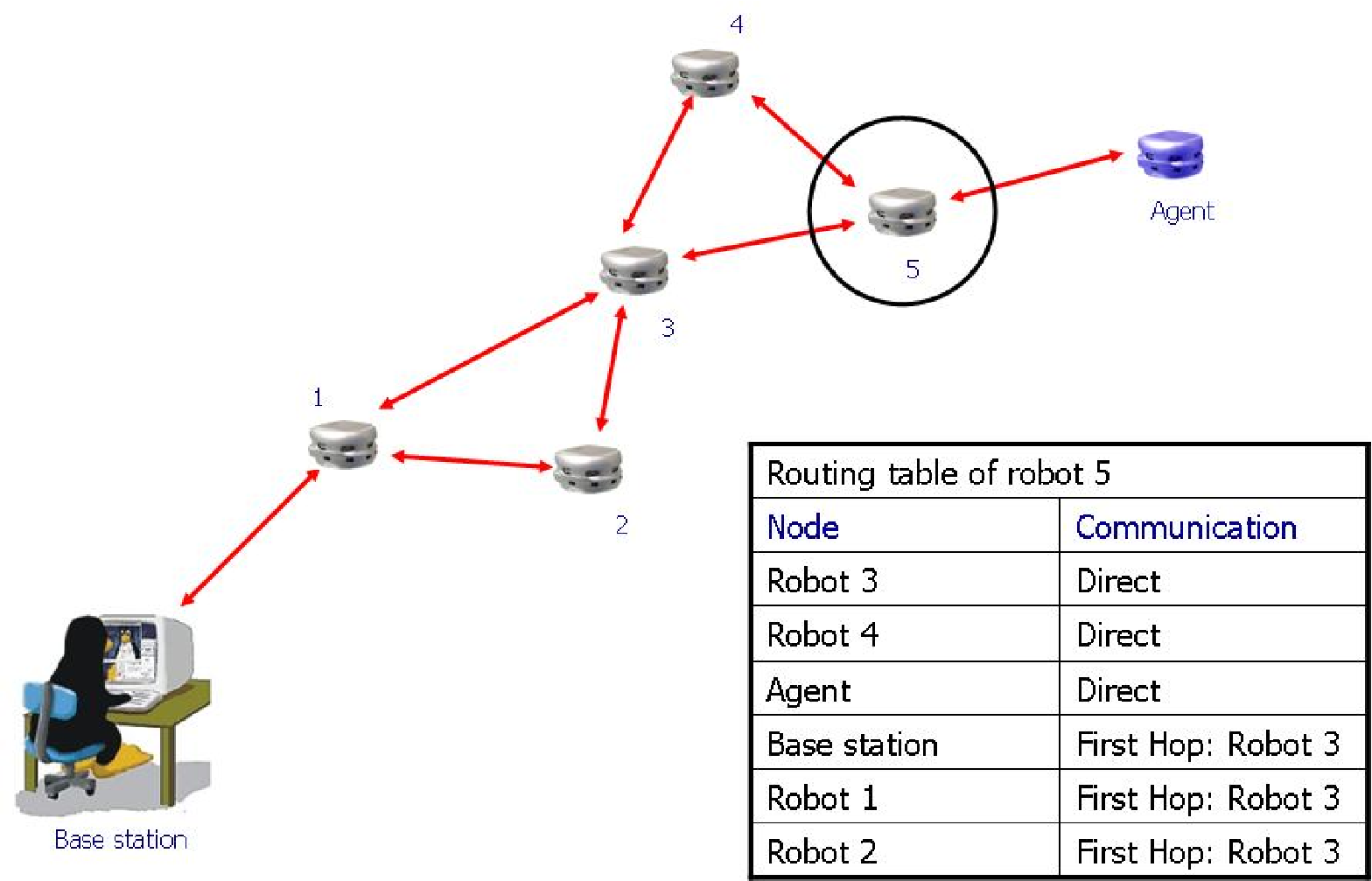}{0pt}{Routing table available to one of the robots in an assigned configuration.}{fig:routing}
\end{psfrags}

The forwarding strategy allows the robots to dynamically realize if they belong to the path from the lead agent to the base or not, and to adopt different motion strategies accordingly. Robots belonging to the path are mainly requested to keep connectivity with the previous and successor robots along the path simultaneously allowing it to stretch; moreover, they have to recover global connectivity if it gets lost. Robots not belonging to the path are required to remain connected to the network and place themselves so that they can become part of the path whenever required. The motion directives to the robots are generated referring to a behaviour based technique, namely the Null-Space based Behavioural (NSB) control, that composes a set of prioritized task functions describing the elementary behaviours.

\begin{psfrags}
	\psfrag{base station}[][]{base station}
	\psfrag{mobile antennas}[][]{mobile antennas}
	\psfrag{agent}[][]{agent}
	\mypsfrag{6}{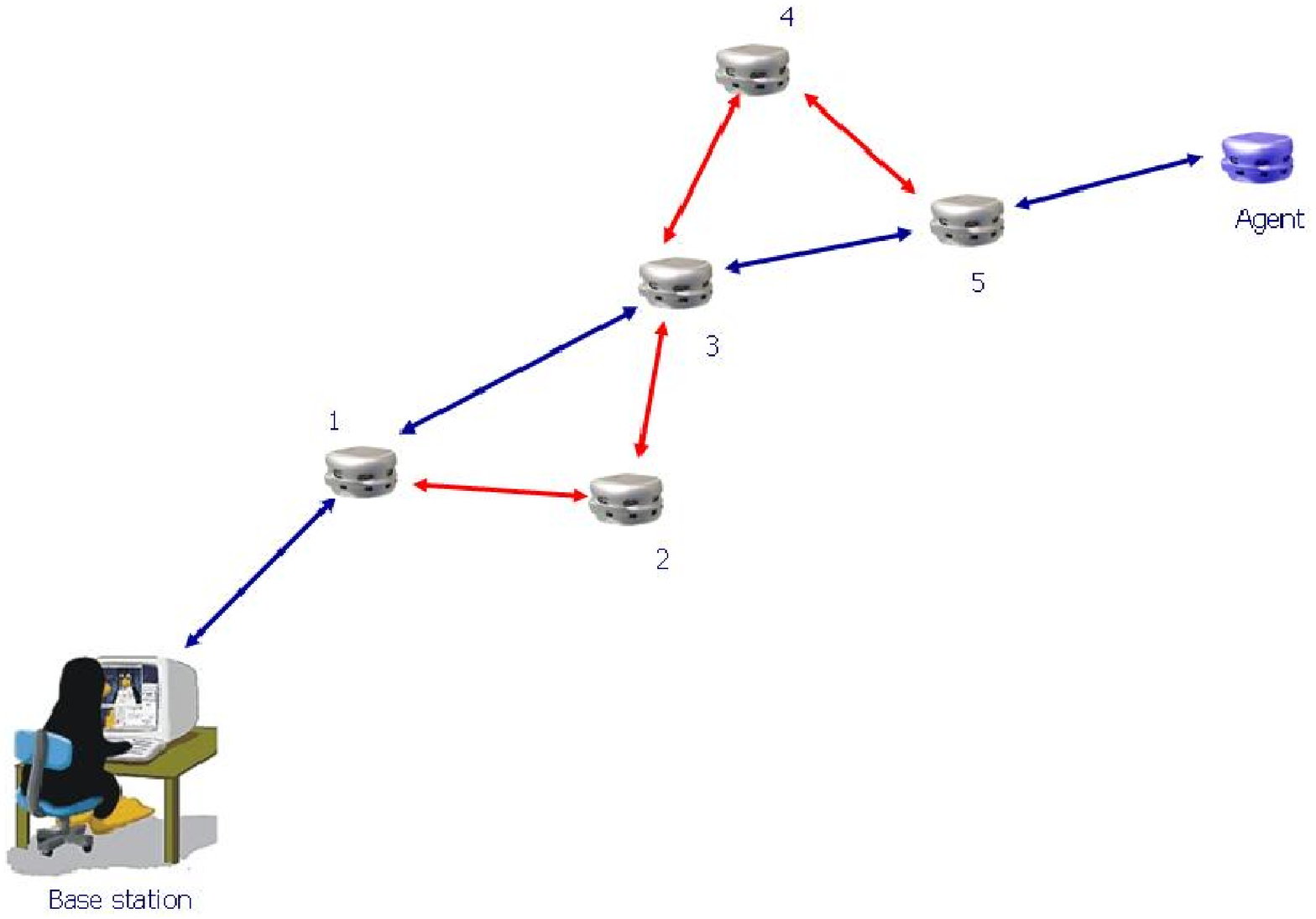}{0pt}{Communication path from agent to base station.}{fig:path}
\end{psfrags}

\section{Null-Space based Behavioral control}\label{sect:control}
The NSB is a behaviour-based approach for robotic systems developed by some of the authors of this paper. The main idea of the approach is to describe the mission through a set of elementary behaviours, to define a task function for each behaviour and to use a projection mechanism (based on the null-space projection matrix) to compose them following their priority order. The behaviours are combined so that the lower priority behaviours do not affect the higher priority ones. The null-space projection matrix of a behaviour filters out the velocity components of the lower priority ones that would affect its functionality. A detailed description of this approach extends beyond the scope of this paper, but can be found in~\cite{AntArrChi_RAM08,AntArrChi_TCST09,AntArrChi_ISR2008}. In the following, we briefly recall the main methodology of the approach and the task functions realized for the specific mission.

Lets define the position of the $j^{th}$ robot as $\bfp_j = \mymatrix{x_j & y_j}\t$ and the generic task variable to be controlled as $\bfsigma\in\Re^m$ a function of the position:
\begin{equation}
	\bfsigma = \bff(\bfp_j)
\end{equation}
The corresponding differential relationship is:
\begin{equation}
	\dot{\bfsigma} = \frac{\partial \bff(\bfp_j)}{\partial \bfp_j}\bfv_j = \bfJ(\bfp_j)\bfv_j
\end{equation}
where $\bfJ\in\Re^{m\times2}$ is the robot configuration-dependent task Jacobian matrix and $\bfv_j \in\Re^2$ is the robot velocity.

The motion directives to the robot, i.e. velocity reference commands, are elaborated as:
\begin{equation}
	\bfv_{d,j} = \bfJ^{\dagger}(\dot{\bfsigma}_d + \bfLambda\tilde{\bfsigma})
\end{equation}
where $\bfJ^{\dagger}=\bfJ\t(\bfJ\bfJ\t)^{-1}$, $\bfLambda$ is a suitable constant positive-definite matrix of gains and $\tilde{\bfsigma} = \bfsigma_d - \bfsigma$ is the task error.

Let us consider the mission for the $j^{th}$ robot composed by multiple elementary tasks. Using the subscript $i$ referring to the $i^{th}$ task quantities for the $j^{th}$ robot, on the analogy of the above equation, the $i^{th}$ task velocity is computed as
\begin{equation}
	\bfv_i = \bfJ^{\dagger}_i(\dot{\bfsigma}_{i,d} + \bfLambda_i\tilde{\bfsigma}_i).
\end{equation}

If the subscript $i$ also denotes the degree of priority of the task with, e.g., Task 1 being the one with highest-priority, the final motion command to the robot is modified into:
\begin{equation}
    \bfv_{f,j} = \bfv_1 + \bfN_{1,1}\bfv_2 + \bfN_{1,2}  \bfv_3,
 \label{eq:srCLIK}
\end{equation}
where $\bfN_{1,k}$ is the projection matrix into the null-space of the task 1 to $k$. In particular, defining $\bfJ_{1,k}$ as
\begin{equation}
    \bfJ_{1,k} = \left[\begin{array}{c} \bfJ_1 \\ \bfJ_2 \\ \vdots\\ \bfJ_k \end{array} \right],
    \label{Jacobiano-composto}
\end{equation}
the null-space projection matrix $\bfN_{1,k}$ is elaborated as \begin{equation} \bfN_{1,k} = \left(\bfI - \bfJ_{1,k}^\dag\bfJ_{1,k}\right).\end{equation}
where $\bfI$ is the identity matrix of proper dimensions. Remarkably, above eq. has a nice geometrical interpretation. Each task velocity is computed as if it were acting alone; then, before adding its contribution to the vehicle velocity, a lower-priority task is projected onto the null space of the immediately higher priority tasks so as to remove those velocity components that would conflict with it.

\section{Connectivity maintenance task functions}
In order to define specific task functions for the connectivity maintenance problem, we assume that the communication capabilities are mainly effected by the relative distances among the network nodes, i.e., the devices can communicate if their relative distance is lower than a certain threshold. In future research, we will consider more realistic models for the wireless channel and we will use specific metrics for e.g., Received Signal Strength Indicator (RSSI) or Packet Loss Percentage; moreover, we will consider effects of the communication environment (as multi-path, noise, and attenuation of signal strength if the devices are not in line of sight). Here, the considered task functions are:
\begin{itemize}
\item  \emph{Distance from a point}. This task function moves the robot at a certain distance $D_d$ from a point $\bfp_1$. Therefore the task function is defined as
\begin{equation}
	\sigma_p = \norm{\bfp-\bfp_1} \in \Re,
\end{equation}
and $\sigma_{p,d} = D_d$. Then, it holds
\begin{equation}
	\bfJ_d = \hat{\bfr}\t \in \Re^{1\times2}
\end{equation}
where $\hat{\bfr} = \frac{\bfp-\bfp_1}{\norm{\bfp-\bfp_1}}$ is the unit vector aligned with the point-to-robot direction. Therefore, the task output is a velocity, in the robot-to-point direction, that keeps the robot at an assigned distance to it.
\item \emph{Equal distance from two points}. This task function moves the robot so as to keep equal distance from two points $\bfp_1, \bfp_2$. 
    The task function is defined as
	$$
		\sigma_e = \norm{\bfp-\bfp_1}^2- \norm{\bfp-\bfp_2}^2 \in \Re,
	$$
the corresponding Jacobian is
	$$
		\bfJ_e = 2\mymatrix{(x_2-x_1) & (y_2-y_1)}
	$$
and the desired value of the task function is $\sigma_{e,d}=0$.
\item  \emph{Move to Goal}.  The move-to-goal task function moves the robot toward an assigned goal:
 \begin{equation}
	\bfsigma_g = \bfp \in \Re^2.
\end{equation}
The desired values  is $\bfsigma_{g,d} = \bfp_{g}$ where $\bfp_{g}$ are the coordinates of the goal. Since $\bfJ_g= \bfJ_g^{\dagger}= \bfI \in \Re^{2\times2}$, the output velocity is given by
\begin{equation}
	\bfv_g = \bfLambda_g(\bfp_g - \bfp)
\end{equation}
a vector in the goal direction proportional to the distance from it.
\item \emph{Obstacles and collision avoidance}. In presence of a point obstacle i.e. $\bfp_o$ in the advancing direction, the aim of the task is to keep the robot at a safe distance from it. Therefore the task function is similar to the \emph{Distance from a point} with difference that it is activated only if the relative distance is under a threshold and the robot is moving in the obstacle direction. The same task function can be used also in presence of a linear obstacle with the difference that the reference point $\bfp_o$ is the point of the segment closest to $\bfp$. For reference ~\cite{ACM_icia12}.
\end{itemize}

A proper combination of such task functions allows to manage the robots to achieve the described mission. In particular, obstacle avoidance, when active, is always the higher priority task since it is needed to preserve robot integrity; however, it is activated only in proximity of obstacles or other neighbouring robots detected by laser range finer or received over wireless link. The other task functions are activated depending on the state of the robots:\\
\emph{Robots belonging to the path} are commanded to keep equal distance form their predecessor and successor nodes; if they loose the connectivity with their predecessor (i.e. the one hop node towards the base) they have to move towards their last known position.
    When the distance from predecessor/successor node is greater than a certain percentage of $r_{\max}$ (the maximum communication range of the device in free space), the robot sends explicit help requests to neighbouring robots that do not belong to the path (if any) and asks one of them to get placed in the middle point between itself and its predecessor/successor node; in this way we facilitate the insertion of other nodes in the path.\\
\emph{Robots not belonging to the path} should stay connected to the network and remain available to join the path if requested. A first solution could be to keep the same distance from the base station and the first node of the route; alternative solution could be to keep equal distance from the first hop towards the base and the first hop towards the agent.

The description above has been coded in a Finite State System activating the proper behaviours within the framework of the NSB coordination strategy~\cite{AntArrChi_ISR2008}.

\section{Experiments}
\subsection{Experimental platform}
The experimental validation has been made with a team of five Khepera~III mobile robots, by K-Team corporation, that are small size ($12$\ts cm diameter) differential drive mobile robots equipped with several proximity sensors ($11$ IR sensors and $5$ ultrasound sensors). The robots also have an extension board, namely the Korebot II, an embedded platform based on the gumstix Verdex PRO with a Marvell PXA270 XScale processor~@~$600$\ts MHz, with $128$\ts MB~RAM and $32$\ts MB Flash memories. The Korebot II runs on standard embedded Linux o.s. (Angstrom distribution, kernel~2.6) and we used a specific C-library of basic functions developed by the Distributed Intelligent Systems and Algorithms Laboratory, EPFL.

\mypsfrag{7}{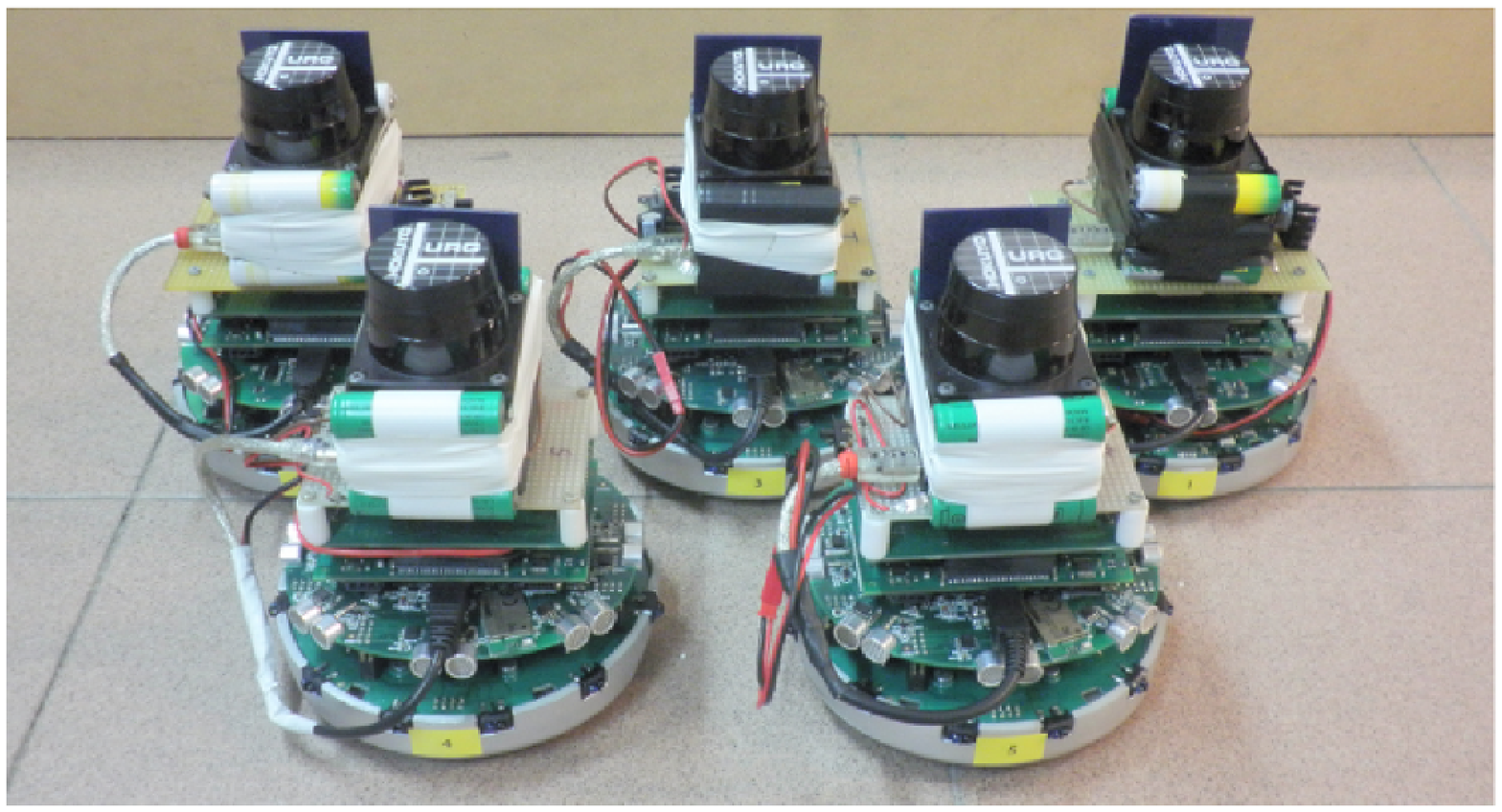}{0pt}{Mobile robots Khepera~III equipped with horizontal laser scanner.}{fig:setup}

To improve the robots' perception we equipped them with Hokuyo Laser Range Finder (LRF) URG-04LX-UG01. The LRF is connected to the Korebot II through USB and power supply is provided by external batteries. The Hokuyo LRF is apt in terms of weight, power consumption, range and accuracy, to properly fit with Khepera III robots and help in navigation in a structured environment. Each robot is able to localize itself w.r.t. a known map by using a line-feature-based Extended Kalman Filter presented in~\cite{ACM_icia12}.

The Korebot board also allows robot-robot and, eventually, robot-pc communication, when embedded with an IEEE~802.11 wireless card. In particular, the robots communicate through a wireless ad-hoc network built with the aim to perform completely distributed experiments. For routing and multi-hop communication we used specific linux functions as {\tt ip\_forwarding} and a routing protocol via {\tt olsrd} daemon. Specifically, olsr daemon is an implementation of the Optimized Link State Routing protocol that allows mesh routing for any network equipment. This protocol runs on any wifi card that supports ad-hoc mode, and it is widely used and well tested. The protocol dynamically generates the routing table of the mesh network; it is pro-active, table driven and utilizes a technique called multi-point relaying for optimized message flooding. Further details on the olsrd implementation go beyond the scope of this paper, while it is worth noticing that the olsrd dynamically generates the routing tables as described in Sect.~\ref{sec:connectivity}.

It is worth noticing that the output of NSB control is a linear velocity command considering the robot as a spatial point; thus, to use it to control a non-holonomic vehicle, as the Khepera  III, the NSB output velocities are passed to a Low-Level control that elaborates angular and advancing velocities and, using odometric model parameters, converts them to left/right wheel velocity commands. In the implementation of the algorithm, the resulting velocity are saturated to feed the system with limited amplitude signals.

\subsection{Experimental results}
In this section we present results of connectivity maintenance experiments performed in an indoor environment (the corridor of our institution). 
 It is worth noticing that, despite the small size of the robots, the dimension of the area is of the order of $90$ m; moreover, since point-to-point communication range is of the order of $20-25$ m, to navigate in the complete environment keeping the communication with a base station (at: $[1.3, -1.5]$ m) multi-hop communication is required.

\begin{figure*}
\begin{center}
\begin{minipage}[c]{4truecm}
\centering
\includegraphics[width=3truecm]{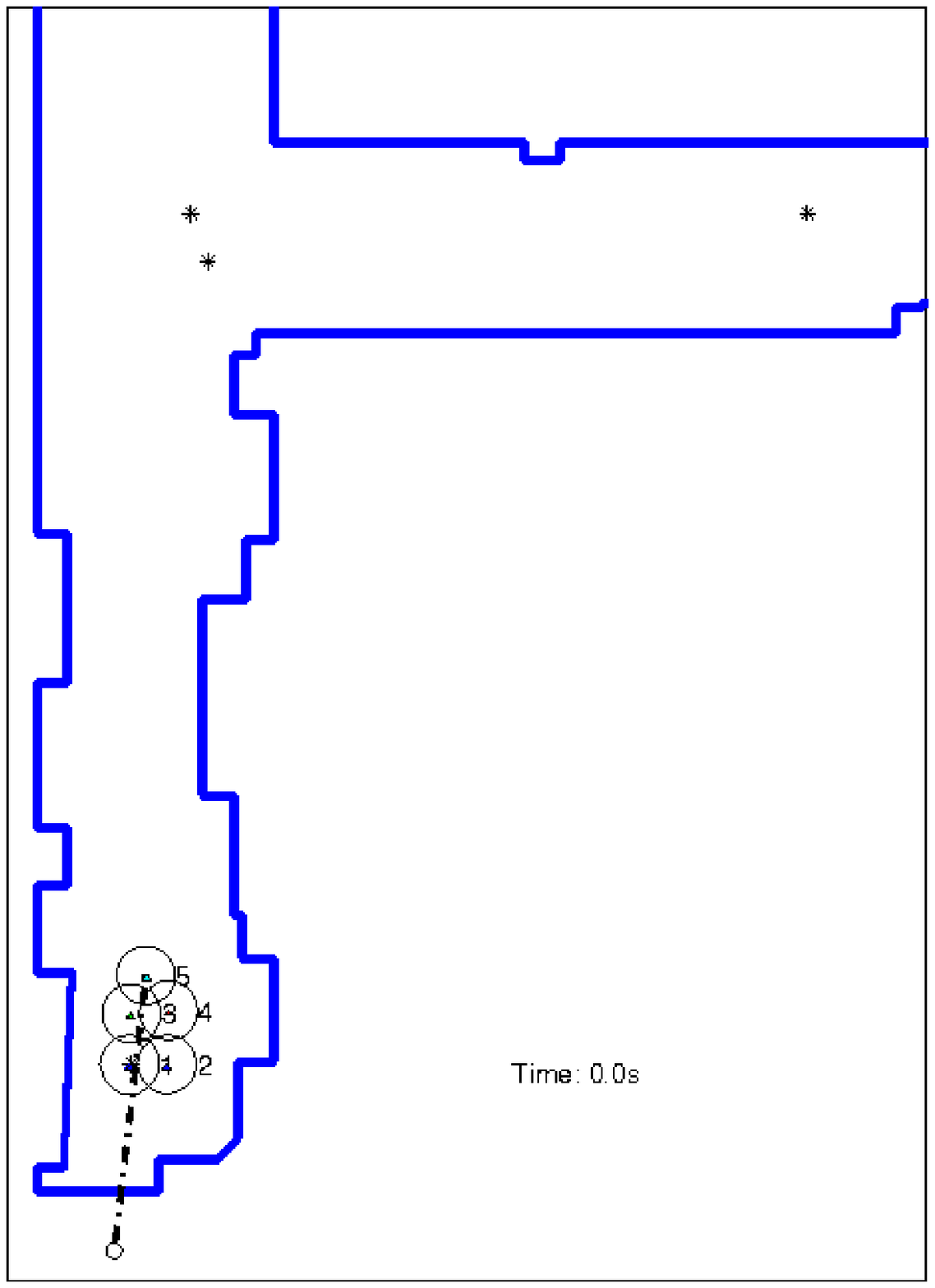}
\end{minipage}%
\begin{minipage}[c]{4truecm}
\centering
\includegraphics[width=3truecm]{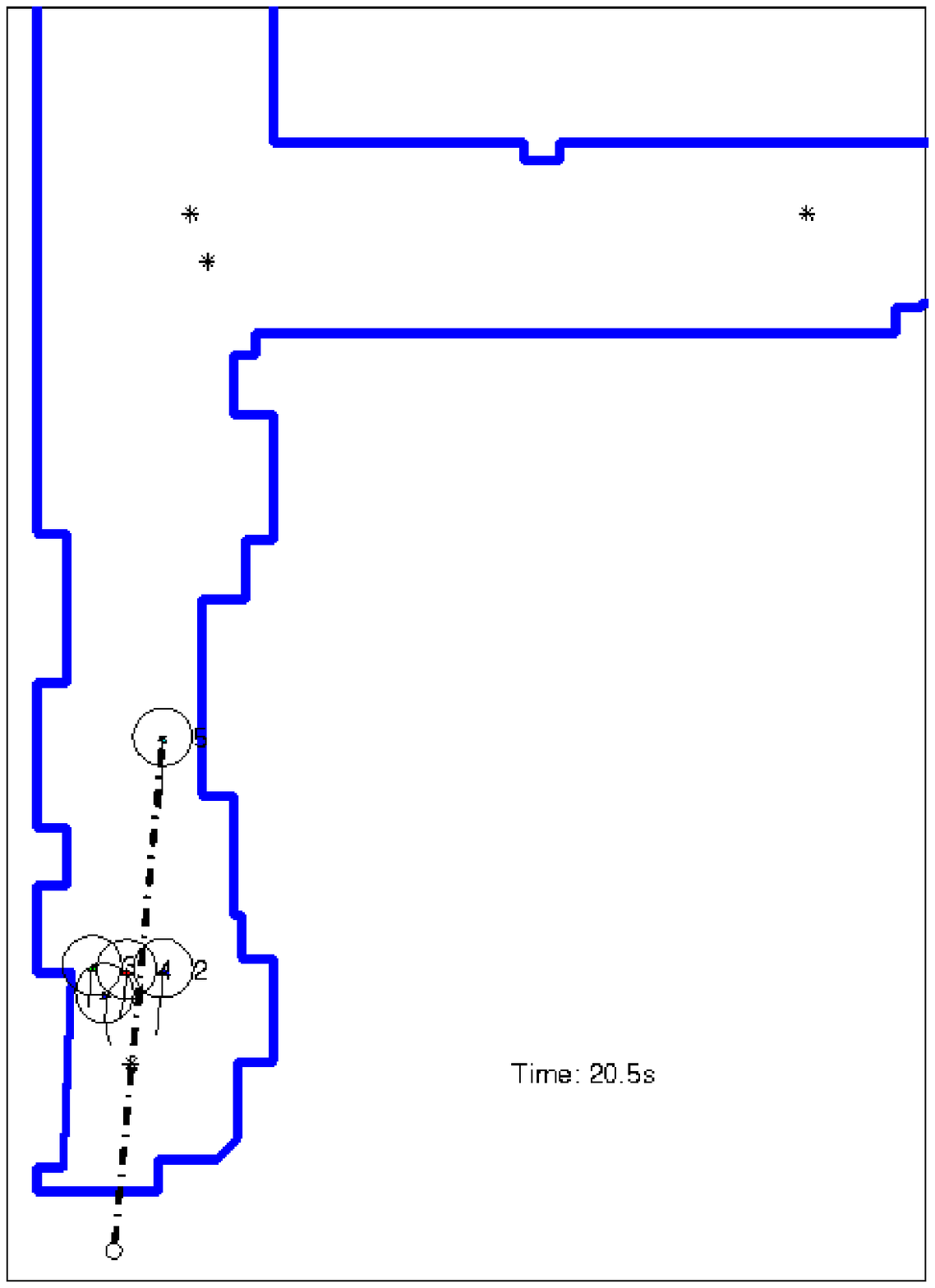}
\end{minipage}
\begin{minipage}[c]{8truecm}
\centering
\includegraphics[width=6.truecm]{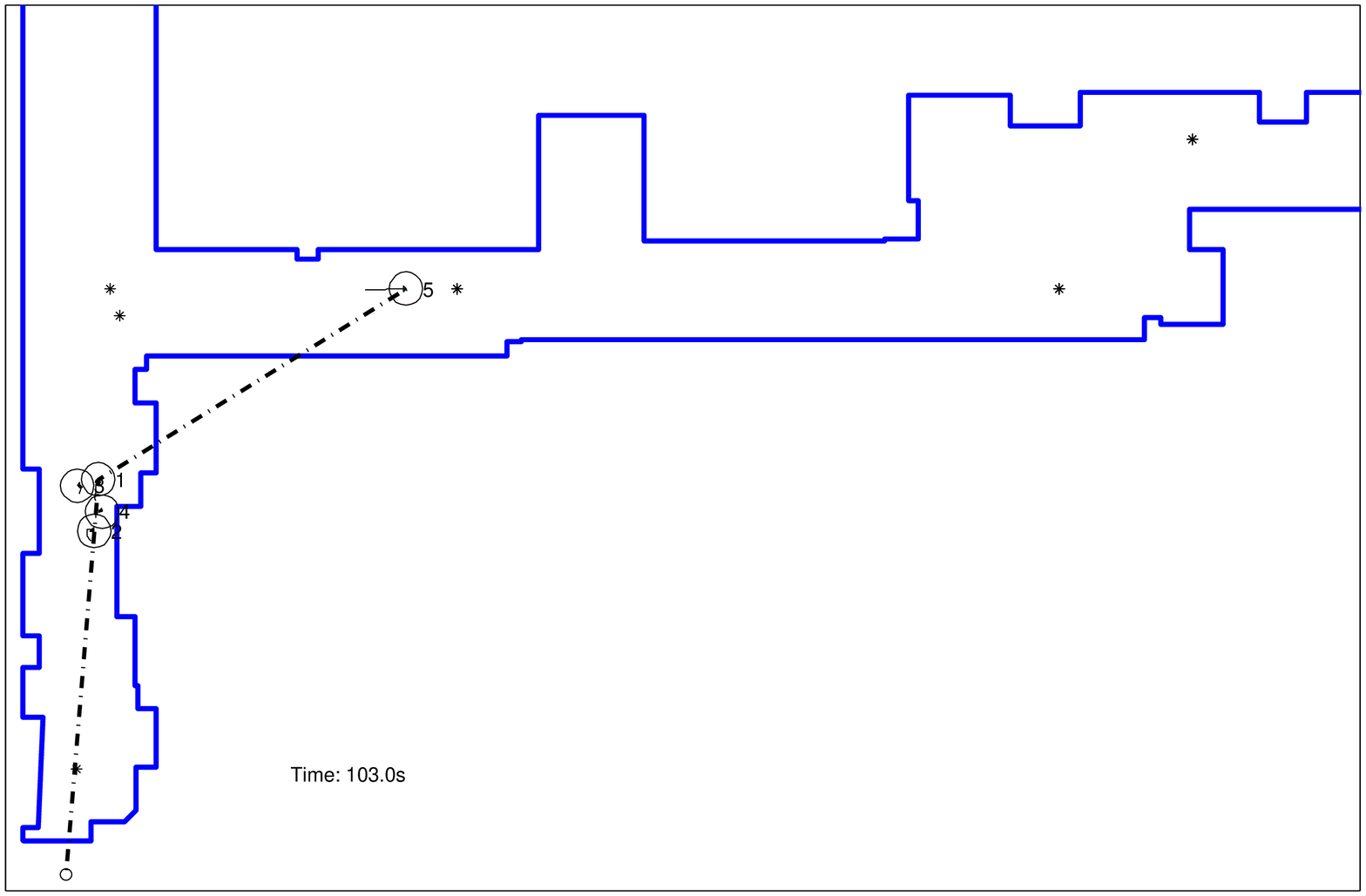}
\end{minipage}\\
\begin{minipage}[c]{8truecm}
\centering
\includegraphics[width=7truecm]{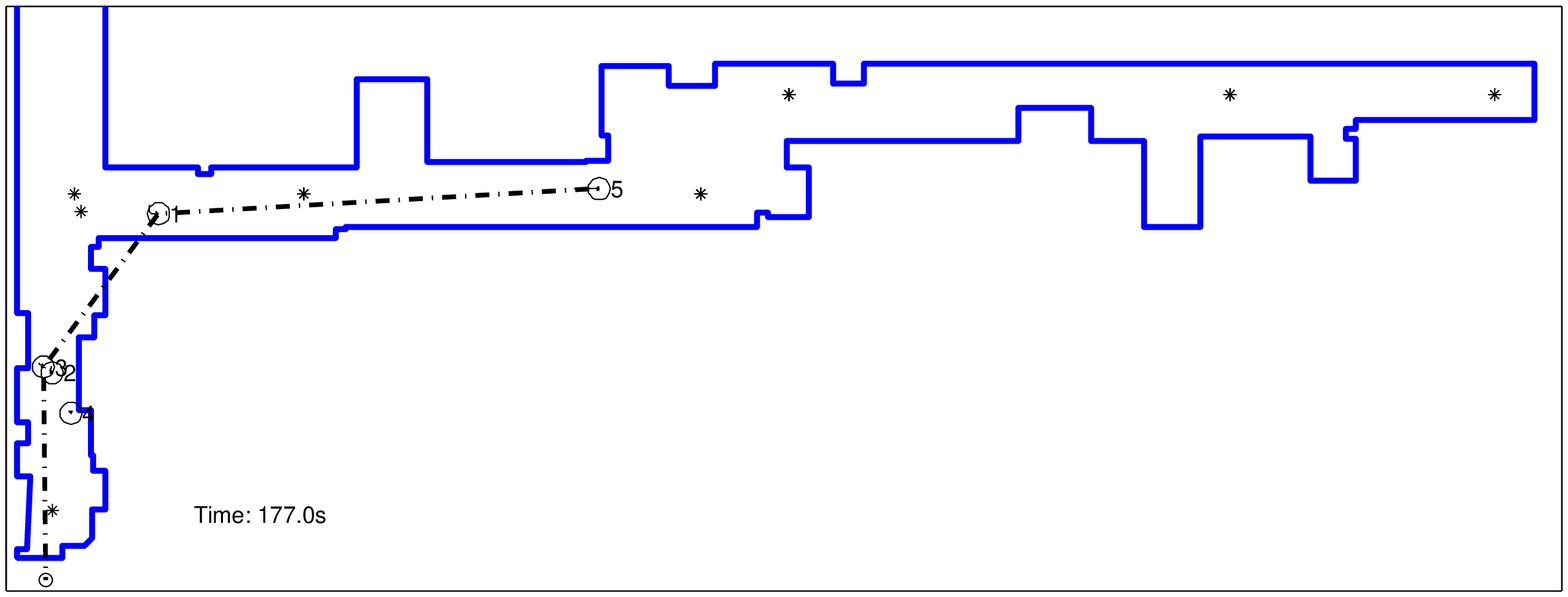}
\end{minipage}
\begin{minipage}[c]{8truecm}
\centering
\includegraphics[width=7truecm]{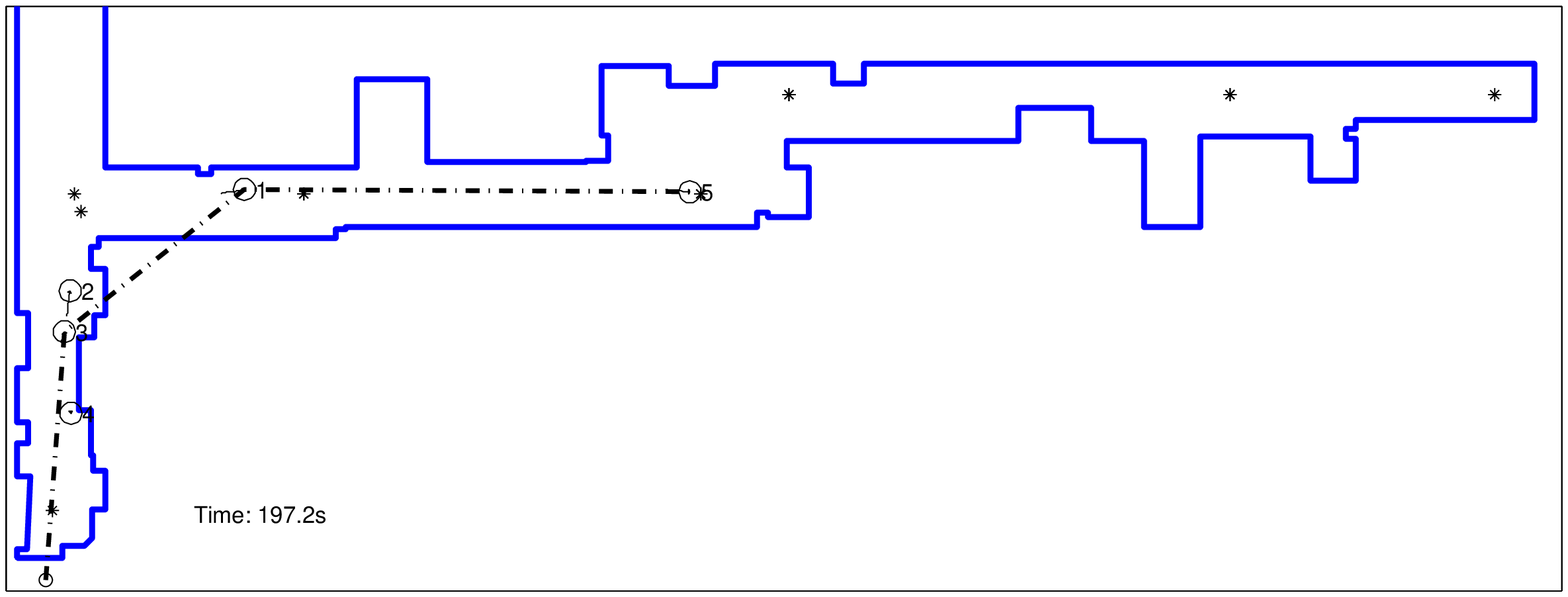}
\end{minipage}\\
\begin{minipage}[c]{8truecm}
\centering
\includegraphics[width=7truecm]{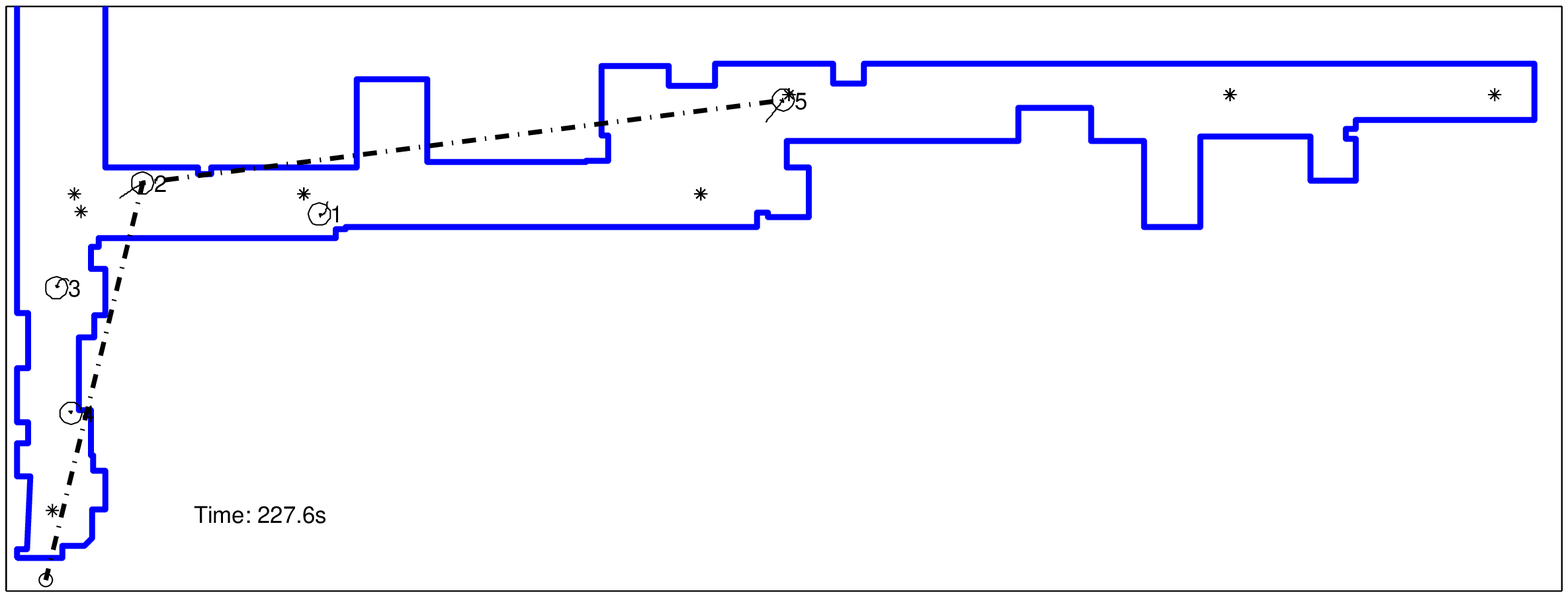}
\end{minipage}
\begin{minipage}[c]{8truecm}
\centering
\includegraphics[width=7truecm]{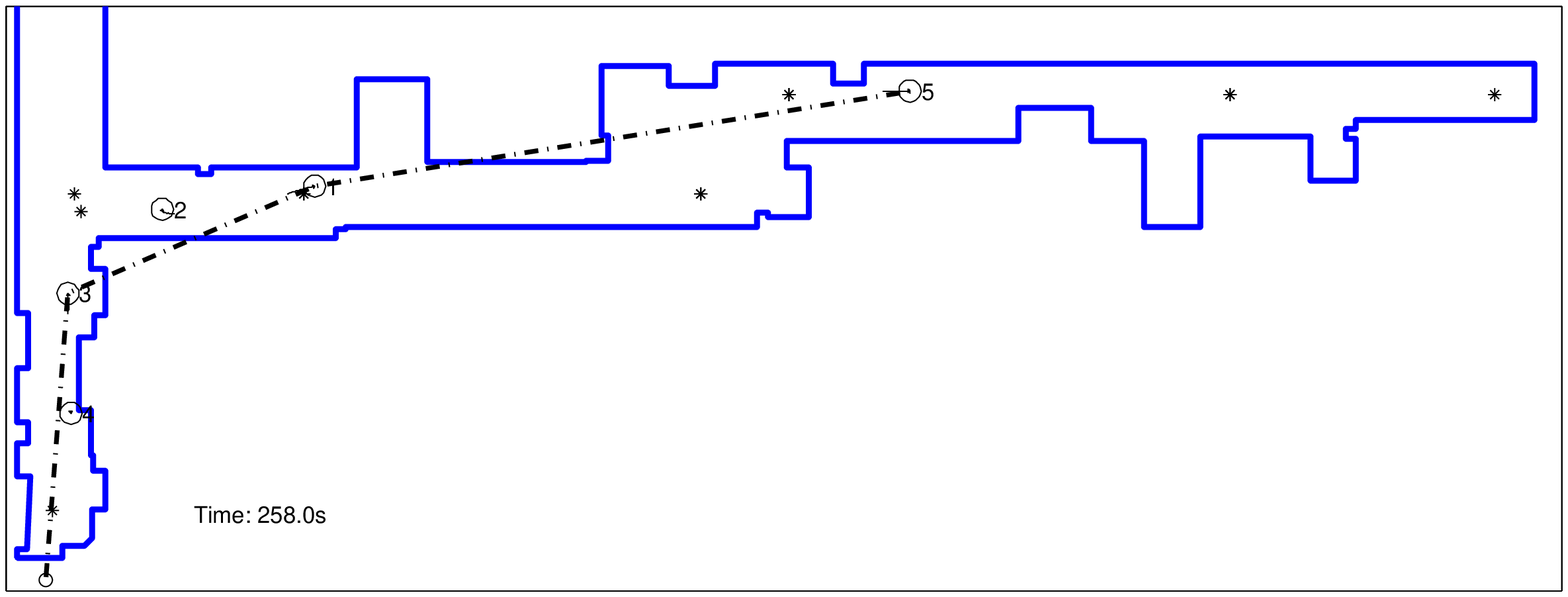}
\end{minipage}\\
\begin{minipage}[c]{8truecm}
\centering
\includegraphics[width=7truecm]{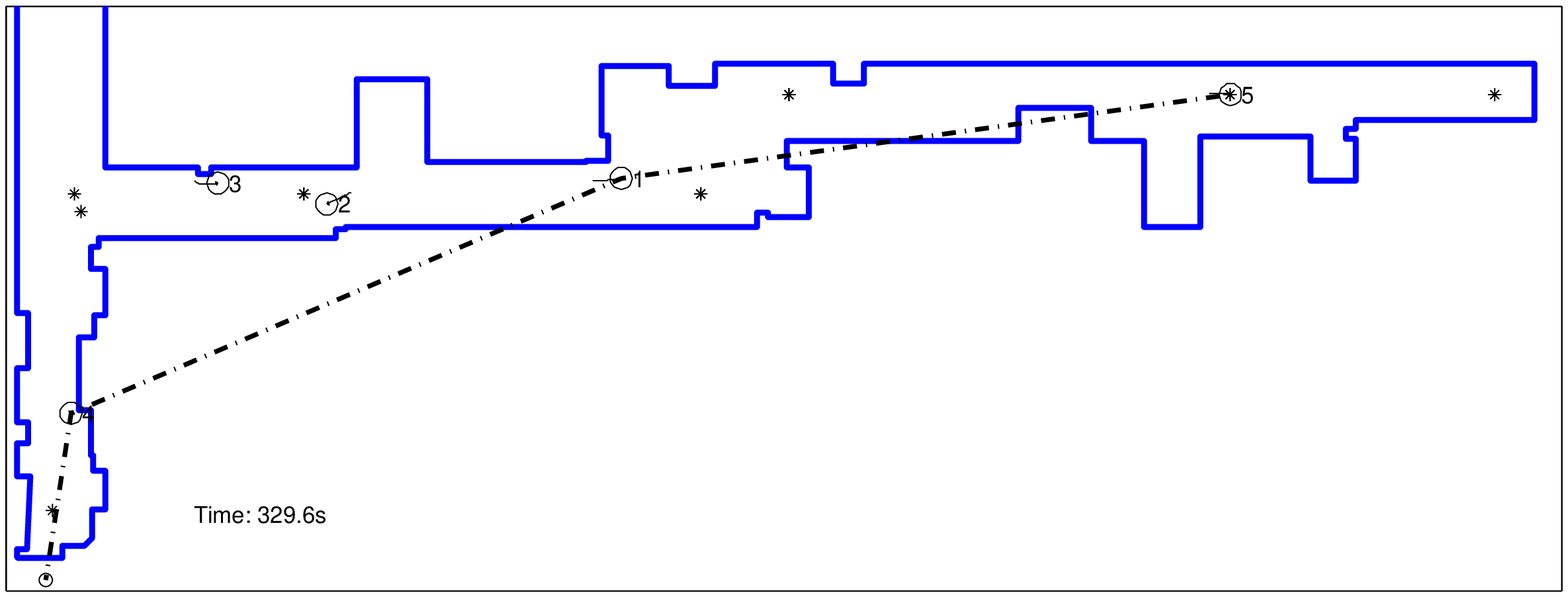}
\end{minipage}
\begin{minipage}[c]{8truecm}
\centering
\includegraphics[width=7truecm]{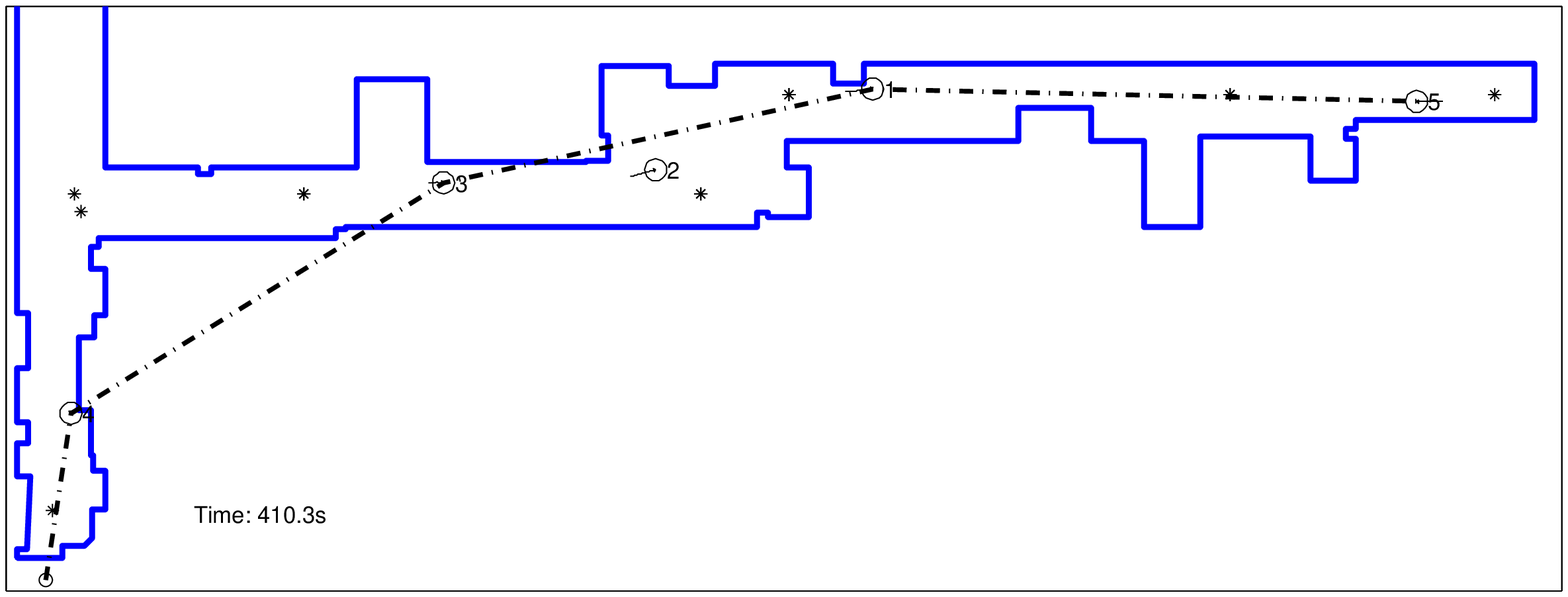}
\end{minipage}
\end{center}
\caption{Paths driven by the five robots during the robotic MANET experiment.}
          \label{fig:corridor}
\end{figure*}


Five robots were used for the experiment. One of them (the agent) was commanded to follow a set of way points spread in the complete area; the others were used as support robots to allow multi-hop relay communication. The motion strategy of the agent, as well as its way points, are unknown to the support robots; the only capability of the agent is to halt and communicate its position once it looses connectivity with the base (direct or multi-hop) and also to come back towards the last known predecessor position if the connectivity is not re-established in a certain amount of time.

Figure~\ref{fig:corridor} shows multiple snapshots of the mission execution elaborated by post-processing robots data from the experiment, in the form of localization estimates and routing tables. From experimental evidence, the maximum range of direct communication achievable with the fixed base and one moving robot is effected by both relative distance and line of sight; in fact, as soon as the agent crosses the first intersection, the communication performance with the base quickly decreases and eventually connectivity is lost. The olsrd protocol requires a few seconds to find the multi-hop connection and incorporate one of the support robots in the path. Once a support robot is included in the path, it tries to keep equal distance between the agent and the base, while the other robots do the same but between the inserted support robot and the base.
It is worth noting that the autonomous agent was finally able to reach the last way point as far as almost~$100$\ts m from the base, and the team of robots was able to dynamically configure in order to provide a proper communication support to the agent. After reaching the last way point, the agent tracks back to its starting position following the same set of way points.

\begin{figure}
\psfrag{t[s]}[][]{\small t[s]}
\psfrag{pl}[][]{\small Packet loss}
\psfrag{0}[][]{\small 0}
\psfrag{5}[][]{\small 5}
\psfrag{10}[][]{\small 10}
\psfrag{100}[][]{\small 100}
\psfrag{200}[][]{\small 200}
\psfrag{300}[][]{\small 300}
\psfrag{400}[][]{\small 400}
        \begin{center}
              \includegraphics[width=8truecm]{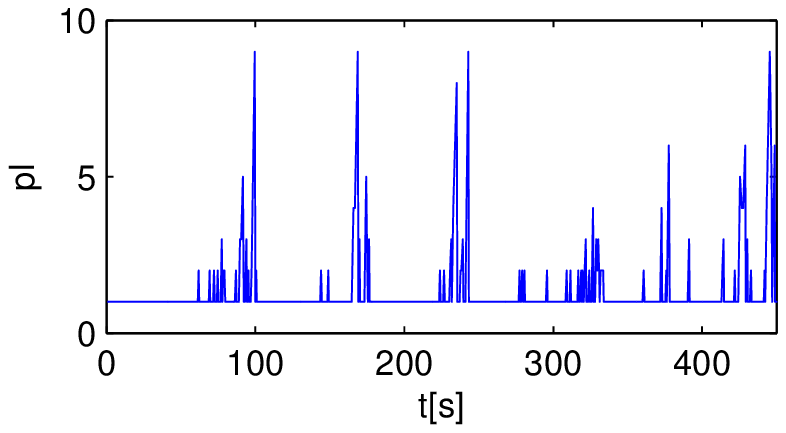}
              \vspace{-10pt}
           \caption{Packet loss during multi-hop transmission between the last robot and the base station.}
           \label{fig:plost}
        \end{center}
        \vspace{-12pt}
        \end{figure}

During the experiment, the agent was constantly required to send data, eventually via multi-hop, to the base station (via UDP/IP protocol). Figure~\ref{fig:plost} shows a plot of the packets lost during the mission generated by checking the identification numbers of the packets received from the base station. It is worth noticing that the global number of packets lost is very limited, and that the loss mainly occurred when a couple of nodes on the path were close to the maximum communication range and the routing protocol was not promptly able to find a new configuration.
	
The maximum speed for the agent and the support robots was set to~$0.2$\ts m/s; the robots' speed is bounded by the limits of the localization algorithm running on board that makes use of laser measurements and it was not lowered for this experiment. The duration of the complete experiment was of$~700$\ts s. 
The video at the following address shows an animation, from experimental data, of the team motion \\http://webuser.unicas.it/lai/robotica/video/manetexp.mp4




\section{Conclusions}

In this work we have presented a distributed motion control strategy for a team of mobile robots to offer multi-hop connectivity between a fixed base station and an autonomous agent. The proposed strategy uses a behaviour-based approach and a dynamic task selection policy to generate motion references for the robots using local information from on-board sensor, communication with neighbours and information from routing tables. The proposed strategy has been successfully experimentally tested in large indoor environment using a team of five mobile robots realizing the robotic Mobile Ad-hoc NETwork.

Future work will concern the refinement of the control strategy to facilitate the introduction of nodes in the communication path between agent and base station. Moreover, we will consider to design {\it complex\/} paths in maze-like environment, i.e., paths that are not driven by a gradient-like policy but require more elaborated strategies to be computed. Finally, we will consider communication performance metrics as Radio Signal Strength Indication or Packet Loss Percentage to overcome distance based task functions.

\section*{Acknowledgments}
The research leading to these results has received funding from the Italian Government, under Grant FIRB - Futuro in ricerca 2008 n.~RBFR08QWUV (NECTAR project).

\bibliography{biblio_unicas}
\end{document}